# A Proposal of Automatic Error Correction in Text

Wulfrano A. Luna-Ramírez, Carlos R. Jaimez-González

Departamento de Tecnologías de la Información, Universidad Autónoma Metropolitana – Cuajimalpa, Av. Constituyentes No. 1054, Col. Lomas Altas, C.P. 11950, México, D.F.
{wluna,cjaimez}@correo.cua.uam.mx

**Abstract.** The great amount of information that can be stored in electronic media is growing up daily. Many of them is got mainly by typing, such as the huge of information obtained from web 2.0 sites; or scaned and processing by an Optical Character Recognition software, like the texts of libraries and goverment offices. Both processes introduce error in texts, so it is difficult to use the data for other purposes than just to read it, i.e. the processing of those texts by other applications like e-learning, learning of languages, electronic tutorials, data minning, information retrieval and even more specialized systems such as tiflologic software, specifically blinded people-oriented applications like automatic reading, where the text would be error free as possible in order to make easier the text to speech task, and so on. In this paper it is showed an application of automatic recognition and correction of ortographic errors in electronic texts. This task is composed of three stages: a) error detection; b) candidate corrections generation; and c) correction -selection of the best candidate. The proposal is based in part of speech text categorization, word similarity, word diccionaries, statistical measures, morphologic analisys and n-grams based language model of Spanish.

## 1    Introduction

The great amount of electronic texts comes mainly from the internet, where the irruption of social networks and the Web 2.0 has provide us a huge quantity of data in just a little period of time. Another great source of information is the data obtained from the scanning of printed documents by means of an Optical Character Recognition (OCR), used by the libraries and government offices in order to preserve files of historic information. Given this enormous quantity of data, there is a need to apply automatic tools to process, transform and extract this information. Even more, a lot of such text contains typing errors and this makes difficult to keep the processing of such texts by other applications; of course, OCR is another source of errors in texts [1],[2],[3],[6],[7]. So, to retrieve text without orthographic errors is a difficult task.

Additionally, when a software tool is used, the main problem of having text with errors is the difficult to use this information as the input of more specialized and even useful applications, like the wide set of Natural Language Processing (NLP) tools [2],[11]. Some examples of such applications are: data mining (and of course web mining), information searching and retrieval, document categorization, and those systems related with education (learning of foreign languages), phonetic translation, code and text computer aided edition, and aided to people with disabilities: text to speech conversion, blind people software assistance, accessibility applications, etc. [9],[10].

Fortunately, the same knowledge of the NLP field can help to design some methods to cope with errors in texts, like the linguistic knowledge. In this way, according to the linguistic point of view, there are different levels of text treatment or processing, some of them are: a) orthographic and morphologic level: describes the structure and external features of words, i.e. the way the letters are articulated in certain language; b) syntactic level: describes the paragraphs and the phases, i.e. the word organization within a text, and the grammatical categories of words, (commonly called Part of Speech -POS); c) semantic level: fully related with the meaning of text, takes care of the context where phrases and words appear; d) contextual or pragmatic level: describe the specific use of words, locutions, phrases, etc., in a certain situation (the discursive and temporal use) within a domain. On this way, the use of certain linguistic knowledge is an adequate approach to design automatic applications of text like the error correction task [1],[2].

In this work, it is used knowledge of the morphologic and contextual levels, specifically a POS tagger, and different techniques of word matching, additionally a language model based in n-grams is used [10],[11].

Briefly, the method is based on speech text categorization, word similarity, word dictionaries, statistical measures, and n-grams model of language.

In the next sections it is exposed the proposal. The section 2 describes the process of error correction and a brief classification of text errors is given; the section 3 describes the method of error correction; after that, some implementation details, experiments, results and discussion are exposed in section 4; section 5 includes conclusion and future work.

## 2  The Process of Error Correction in Electronic Text

The Error Correction in Electronic Text is performed in three stages [1],[6],[7]:
1. **Error detection**: It is oriented to text revision in order to identify the chain of characters from the text (words and other marks) as part of a given language, typically by means of a dictionary comparison, or a grammatical structure (morphological, syntactical or semantic).
2. **Candidate correction generation**: detects some possibilities of correction to a given error.
3. **Correction**: it is the selection of a specific candidate correction and the substitution of it in the text.

In **interactive error correction** the system corrector performs the first and the second stages in an autonomous way, and left to the user the final stage, where the decision of which is the real correction is taken. In the other hand, the **automatic error correction**, the system does the three stages without the need of the user final decision [1].

As can be inferred, the automatic error correction is necessary to PLN applications where the full processing is performed with no user intervention.

Even more, the correction can be divided in two categories [1]: a) **isolated word error** correction: where the three stages described before are performed just word by word; b) **the contextual error correction**: where can be observed the context of the word, i.e. the phrase where it is used.

On the other hand, there is a brief classification of errors [1,7] that can be founded in text (obtained from humans –typing- or by machines –mainly OCR methods). Some error can derive in linguistic mistakes by the accidental generation of invalid words within the language, or can be match with another valid word but not the correct one. The error classes could be part each other, but in order to understand the nature of error it is useful to try one classification, which is depicted in **Figure 1**: a) typographical: contains wrong characters in the string or word; b) grammatical: violate the articulation rules of a certain language; c) cognitive: the origin of the errors is the lack of knowledge of orthographic rules; d) phonetical: they are wrong representations of a given linguistic utterance (a phonetic chunk of language), those are the worst case of error, because generate a great word malformation and the semantic knowledge implied is high.

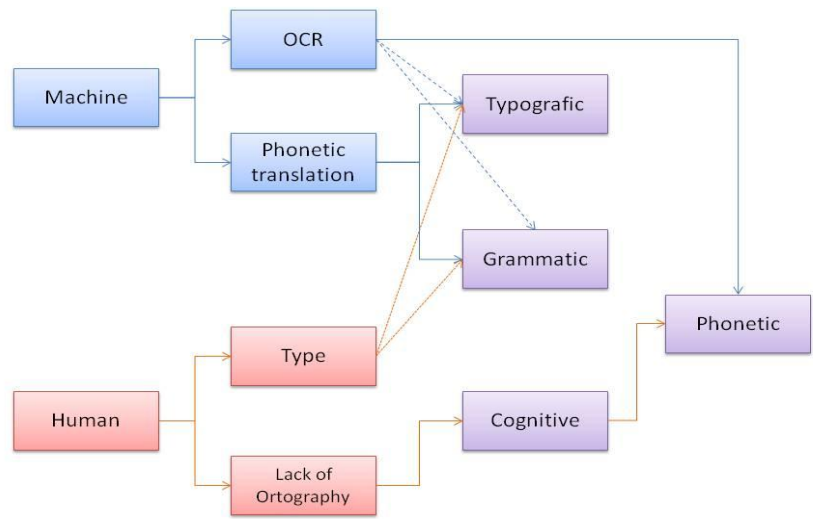

**Figure 1**. A classification of error in text according to their origin: human being or computer systems. One error can originate different cases of linguistic mistake, as is depicted,

for example an OCR error can be derivate in a phonetic error.

From the linguistic knowledge, those different types of errors can be organized accordingly with the language level related: a) morphologic; b) syntactic; c) semantic; d) speech structure; and e) pragmatic.

The types of errors are used to design techniques which can be applied to the stages of error correction in texts. One case of error that has been mainly treated by PLN applications are word error, which are described in the following subsection given the importance for the automatic or semiautomatic error correction methods that had been tried [1].

### 2.1 Word level errors

The level word errors can be treated relatively easy by a computer program, but the propagation of the mistake accordingly with the linguistic levels needs to be coped with techniques which imply knowledge about the other levels of the language. The most successful application use morphological and syntactic knowledge, but the superior levels when are incorporated in them originate domain dependent applications.

The two main sources of electronic texts (human or OCR) introduce certain classes of errors that had been afforded by different algorithms, like the edit distance (called Levenshtein distance) for the morphological errors; and n-gram based comparisons, for the syntactic and (in a very restricted sense) semantic errors [1]:

- **Insertion**: from the addition of one character in the string.
- **Duplication**: the doubling of one character in the string.
- **Deletion**: to omit one character in the string.
- **Substitution**: to replace one character by other.
- **Transposition**: where two characters are swapped.
- **Segmentation**: when two strings are formed from just one word.
- **Union**: when two words originate one string.

The segmentation errors are the most frequent in texts from OCR, on the other hand, the human being commonly produce union errors when typing.

### 2.2 N-grams

In order to verify if a sequence of linguistics elements (words, letters, phrases, etc.) within a text is valid for a given language, it is common use models of languages which represent the constraints of a certain combination of linguistics elements in order to build right structures in language [1],[11],[12].

One model of languages is n-grams, a statistical model of sequences of linguistic elements: typically n word or letters where n >= 2. It is based in the estimation of sequences probability calculated from a text corpus called the **transition probability**, it can give a prediction of the n next element from the n previous according to the frequency with they appear in corpus. N-grams model uses the history based Markov supposition: a linguistic element is affected by the local context, so the previous text affects the future text [1],[11]. Some examples of n-grams or letters and words are showed in **Figure 2**.

| Letters N-gram order | N-gram example |
|---|---|
| monogram | {i,n,f,o,r,m,a,t,i,c,s} |
| bigram | {in,nf,fo,or,rm,ma,at,ti,ic,cs} |
| trigram | {inf,nfo,for,orm,rma,mat,ati,tic,ics} |

| Words N-gram order | N-gram example |
|---|---|
| monogram | {computing,is,not,easy} |
| bigram | {computing,is}{is,not}{not,easy} |
| trigram | {computing,is,not}{is,not,easy} |

**Figure 2**. The first table shows n-gram examples of the word *informatics*, and the second contains n-gram examples of the phrase: *computing is not easy*. Where n = {1,2,3}

N-grams had been used in error detection and the candidate generation stages of the error correction task [1]. In the proposal presented in this paper lexicons of bigram and trigram were used.

In the next sections the proposal of automatic error correction and the implementation are exposed.

## 3 The Automatic Error Correction Proposal

The error correction can be focused to the isolated words or taking account the local context of them. The proposal presented in this paper (POS-Tagged Automatic Error Correction -PAEC) is oriented to automatic error correction of the chain of words in a paragraph, and eventually in the full text [14]. It is based on POS text categorization, word similarity comparison, word dictionaries, statistical measures, and n-grams language models.

Some techniques used in automatic detection, candidate generation and selection of the correction use probability and morphology of words. The proposal adds to those techniques a POS-tagging process for the sake to augmenting the linguistic knowledge in the full process, i.e. uses the morphological and syntactical information present in the text under revision.

The proposal uses the following resources (texts written in Spanish): 1) a plain text corpora (CT), 2) a POS-tagged corpora (CA) with ten grammatical categories defined: verbs, nouns, conjunctions, idioms, articles, adjectives, adverbs, pronouns, interjections and miscellaneous; 3) a text corpora with inserted errors (CE), from a OCR and random deletions, insertions or substitutions; 4) a set of word lexicons: nine of each POS defined, and one of words from all grammatical categories; 5) a n-gram model of language (bigrams and trigrams); and 6) a n-gram model of language of POS-tags, which identify the n-gram occurrence of certain POS-tags (bigrams and trigrams).

The PAEC proposal is showed in **Figure 3**. It is composed of the following modules for to analyze a text:

1. **Preprocessing**: the abbreviation words are expanded, upper case and punctuation symbols are extracted, and the resulting text is separated in sentences and words.
2. **POS-tagging**. The text is POS-tagged in order to identify the grammatical category they belong. The POS-tagger used is an implementation of TBL POS-tagger [4], trained which semiautomatic annotated Spanish text.
3. **Word extraction**: it separates the words and the POS-tagged related, in order to recover the original text, but identifying the grammatical category of words. Produce a list of word for being analyzed.
4. **Contextual error detection**: this task makes a list of possible wrong words in text. It is carried out in two phases:
    a. Seeks each word from the text in their corresponding POS-lexicon.
    b. An analysis of bigrams that can be formed with the words from the text for identifying the abnormal combinations, because it can discover the presence of an error in text.

The process is showed in the **Algorithm 1**, as follows:

```
Input:
    P: List of words to be corrected
    LN: Lexicon of language model (n-grams)
Output:
    PE: List of words identified as errors
Var:
    k: It represents the index of P word it is being re-
    viewed in each iteration.
    P_k: It is the k member of P
    |P|: Total number of words in the list P
    LP: Word lexicon
    pa: It is the word is being reviewed in each iteration
    (P_k)
    ant: previous word of pa
    post: next word of pa
    flag: It indicates if a bigram of pa has been found
    <I>: It is an inserted pseudo-word at begin of each
    line
    Functions:
    add(l,e): add the e element to the list l
    Begin
1.  k = 0
2.  While k < |P|  do
3.    ant = P_k-1
4.    pa = P_k
5.    post = P_k+1
6.    If pa != <I>
7.        If pa not in LP then
8.            add(PE,pa)
```

```
9.         else
10.        If [ant,pa] not in LN then
11.           If ant not in LP then
12.              add(PE,ant)
13.           else
14.              flag = 1
15.        If [pa,post] not in LN  then
16.              If post not in LP then
17.                  add(PE,post)
18.              else
19.                  If post not in PE
20.                      flag = 2
21.        If flag = 2 then
22.              add(PE,pa)
23.        If flag = 1 then
24.              If ant not in PE then
25.                  add(PE,pa)
26.   k = k + 1
27.   Return PE
   End
```

**Algorithm 1**. Contextual Error Detection. This algorithm receives the list of words to be corrected and the lexicons where the words are searching. It generate the words identified as errors, this is the input for the next stages of error correction.

5. **Potential corrections generation**: taking in count the POS-tag of the potential errors detected in the previous modules, it makes a candidate generation of corrections based in n-grams and morphological comparisons (skeleton and edit distance), seeking in the right lexicon according to the grammatical category of words.
6. **Correction**: it selects the best potential correction from the candidate words, this is the responsible for taking the right one. This selection is based in: a) n-gram analysis for identify the most probable combination of those candidates and the local context of word; b) minimum edit distance of words and candidates; c) skeleton comparisons; and d) size of words and word skeletons.
7. **Postprocessing**: once the correction has been done in text, the upper case, abbreviations and punctuation symbols are reintroduced and the final text is saved.

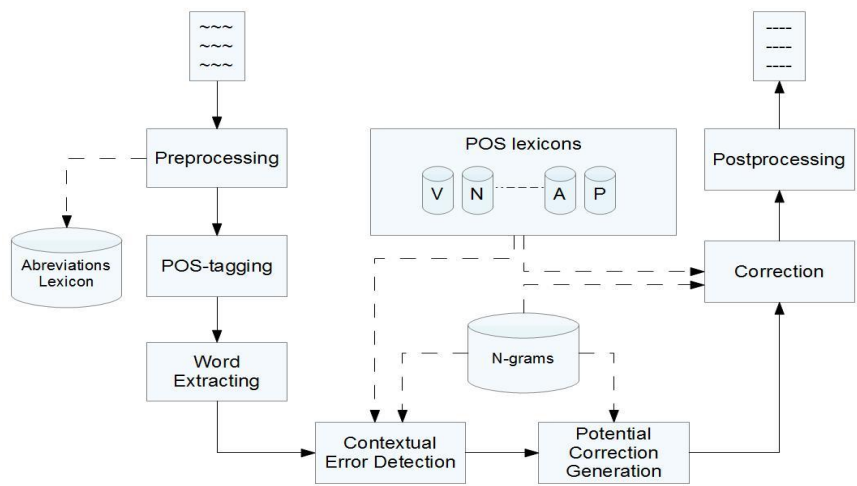

**Figure 3.** The PAEC proposal. The automatic error correction method is based in the identification of POS of the text and the model of language based in n-grams.

In the following section are discussed the implementation matters, and the experiments carried out in order to verify de efficacy of PAEC proposal.

## 4 Implementation, experiments and results

The PAEC proposal was implemented using the PERL programming language, because it is relatively easy to construct text analyzer functions, string matching, and lexicon searching [14].

In order to test the PAEC proposal of error correction an alternative method of correction was implemented Morphological Automatic Error Correction (MAEC). This method does not include a phase of POS-tagging for the sake of represent a more traditional way to do the error correction in text, i.e. it contains less linguistic knowledge, specifically morphologic and syntactic information, so a framework of comparison can be established. The MAEC method is depicted in **Figure 4**.

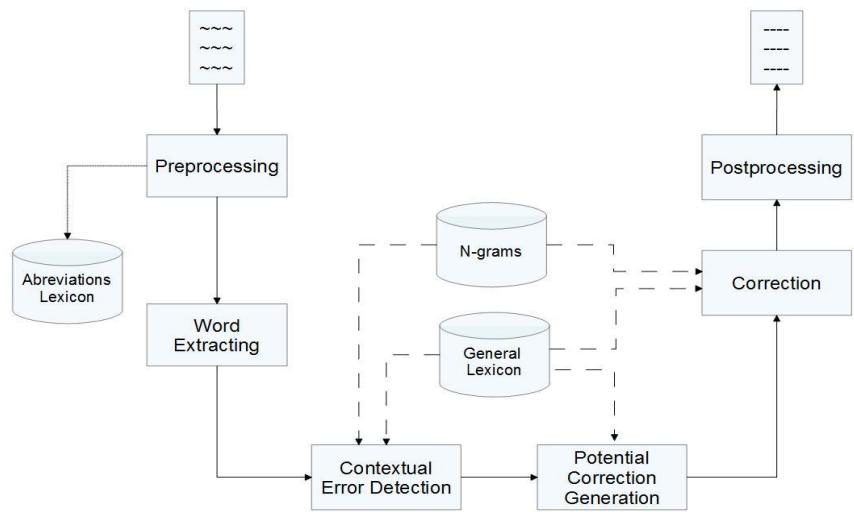

**Figure 4**. The Morphologic Automatic Error Correction (MAEC) method, it uses a n-grams model of language and a General Lexicon for making the three stages of text error correction. It was used for testing the PAEC proposal of error correction.

## 4.1   Data description

As it was mentioned in the previous section, many linguistic resources were used to implement the PAEC and MAEC methods of error correction in texts. They are described in the **Tables 1**, **2**, **3**, **4, 5** and **6**.

**Table 1**. Plain Text Corpus characteristics. The texts were extracted from different Web sites written in Spanish from the following domains: Essay, Politics, Literature and Philosophy.

| Language | Spanish |
|---|---|
| Format | ASCII |
| Files | 497 |
| Text lines | 137,103 |
| Words | 2,877,834 |
| Characters | 18,075,061 |

**Table 2**. The POS-tagging manually annotated text corpus characteristics. This corpus was used for the initial training of the POS-tagger which is an iterative and supervised process, where results in a bigger (automatically) annotated corpus for being used in the error correction method. In the first stage it was used 0.1509% of the plain text corpus.

| Language | Spanish |
|---|---|

| Format | ASCII |
|---|---|
| Files | 1 |
| Text lines | 243 |
| Words | 4,346 |

In the **Table 3** is described the set of testing files, which contains a lot of errors, one set is obtained from an OCR (Hewlett Packard, ScanJet 3200c, their features can be seen in the **Table 4**) process and the other one was generated according to the following process (for simulating typing errors):
1. The file is formatted in order to keep 10 words per line.
2. One word of each line is randomly selected
3. From each word it is randomly selected one character
4. In the position selected is randomly inserted one error (insertion, deletion or substitution)
5. The modified word is reinserted in its corresponding position in the file.

So, it tis gotten one random error each ten lines.

**Table 3**. Text Corpus with errors used for testing the error correction methods. The *Errors 1* column shows the automatically inserted errors. The *Errors 2* refers to the error from the OCR process.

| Texts | Lines | Strings | Characters | Errors 1 | Errors 2 |
|---|---|---|---|---|---|
| 1 | 15 | 79 | 587 | 7 | 5 |
| 2 | 17 | 92 | 634 | 7 | 8 |
| 3 | 16 | 126 | 831 | 13 | 2 |
| 4 | 11 | 147 | 1068 | 14 | 9 |
| 5 | 14 | 150 | 985 | 15 | 4 |
| 6 | 19 | 260 | 1770 | 25 | 28 |
| 7 | 17 | 310 | 2061 | 30 | 10 |
| 8 | 20 | 358 | 2045 | 55 | 52 |
| 9 | 17 | 591 | 3800 | 116 | 29 |
| 10 | 19 | 1153 | 6774 | 41 | 15 |
| **Total** | **165** | **3,266** | **20,555** | **323** | **162** |

**Table 4**. Features of the scanner and the OCR software used to generate the *Error 2* set of files.

| Scanner | HP ScanJet 3200c |
|---|---|
| ORC sofware | HP Precisionscan LT |
| Resolution | 600 x 1200 dpi |
| Format of output | Plain text (.txt) or riched text (.rtf) |

The set of lexicons are showed in the **Table 5**. The General Lexicon is used by the MAEC method, while the other lexicons (POS-based) are used by the proposal PAEC. The contents of each lexicon were taken from the Dictionary of Usual Spanish [12], and the structure of them is given by the orthographic form or each word and their n-grams of letters. Word between one to two letters has just their monogram, but words of four letter or more has either bigram and trigram.

**Table 5**. The features of the lexicons used by the error correction methods. A = words total number. B = larger size of word. C = smaller size of word. D = media of size of words.

| Lexicon | A | B | C | D |
|---|---|---|---|---|
| General | 131,648 | 42 | 1 | 17 |
| Verbs | 10,206 | 30 | 2 | 11 |
| Prepositions | 32 | 13 | 1 | 17 |
| Pronouns | 83 | 8 | 2 | 5 |
| Interjections | 26 | 7 | 2 | 5 |
| Conjunctions | 38 | 20 | 1 | 8 |
| Articles | 9 | 3 | 2 | 3 |
| Adjectives | 5,508 | 26 | 2 | 14 |
| Adverbs | 1,516 | 23 | 2 | 12 |
| Nouns | 14,517 | 25 | 1 | 12 |

**Table 6**. Features of the N-grams based model of language: larger, smaller and media size of just bigram are showed.

| Bigrams | 566,521 |
|---|---|
| Larger frequency | 25,542 |
| smaller frequency | 1 |
| Media frequency | 4 |

### 4.2 Experiment Description

The experiment carried out for testing the MAEC and PAEC methods of error correction in text is composed of two evaluations:

- **Simulation of typing errors** (errors automatically inserted): the purpose was to identify the performance of insertion, deletion and substitution errors.
- **OCR errors in text**: this is a life real set of evaluation, the files comes from an OCR method after being scanned.

Additionally, for the PAEC proposal, one more test was made: manually POS-tagged text was tested in order to obtain an ideal tagging, and verify the efficacy of the method by means to avoid the possibility of getting some deviation because the precision of the automatic POS-tagger.

The following accounts were made in the texts from both methods:

C: corrected words. Real errors detected and corrected properly.

E: detected errors. Total of errors identifies by the methods.

e: non detected errors. Total of errors not identified by the methods.

I: introduced errors. Wrong substituted words due to the methods identify some false errors.

F: false error correction. The candidate correction selected for correcting the error detected was wrong, so the error persist.

o: original errors. They are the original errors presents in text.

On the other hand, the rate or errors was calculated using the following formula [66]:

$$c = p - i / p$$

Where c: rate of error correction; p: words of text; and i: errors of text (wrong words).

### 4.3 Results of experiment

The results of the testing of MAEC method of error correction are showed in the **Table 7** and the results of PAEC method are showed in **Table 8**. The results of the third experiment (manually POS-tagged files) are showed in **Table 9**.

**Table 7.** The results of the automatic error correction based in n-grams and morphologic analysis MAEC. The first row is the result of the typing simulation set of testing, while the second row is the result of the texts from the OCR.

| Set of testing | p | o | i | C | E | e | I | F | c |
|---|---|---|---|---|---|---|---|---|---|
| Errors 1 | 3266 | 323 | 140 | 202 | 311 | 31 | 19 | 90 | 0.957 |
| Errors 2 | 3185 | 162 | 104 | 86 | 180 | 10 | 38 | 56 | 0.967 |

**Table 8.** The results of the automatic error correction based in n-grams and morphologic analysis PAEC. The first row is the result of the typing simulation set of testing, while the second row is the result of the texts from the OCR.

| Set of testing | p | o | i | C | E | e | I | F | C |
|---|---|---|---|---|---|---|---|---|---|
| Errors 1 | 3266 | 323 | 370 | 112 | 474 | 28 | 167 | 180 | 0.886 |
| Errors 2 | 3185 | 162 | 229 | 84 | 304 | 8 | 152 | 70 | 0.928 |

**Table 9.** The results of the automatic error correction based in n-grams and morphologic analysis PAEC. *Errors 3* row is the test performed with manual POS-tagged files that come from the simulated typing errors, while the *Errors 4* are files that come from an OCR process.

| Set of testing | p | o | i | C | E | e | I | F | c |
|---|---|---|---|---|---|---|---|---|---|
| Errors 3 | 3266 | 323 | 133 | 217 | 238 | 26 | 27 | 80 | **0.959** |
| Errors 4 | 3266 | 162 | 74 | 104 | 186 | 8 | 33 | 45 | **0.977** |

### 4.4 Discussion

As can be seen in the results tables, in the experiments on the sets of files *Errors 1* and *Errors 2* the MAEC method is better than PAEC proposal. These results are discussed as follows.

The MAEC method has a high rate of no detected errors (*e*), this affects mainly to the selection of errors because it introduces new errors (*I*) originated by errors that originate valid Spanish words, as a consequence the wrong word is not identified as error but belongs to a n-gram with low frequency; so, another word would be treated as an error.

The most mismatched words are words of one or two letters: y, a, e, la, el, lo, no; because when a deletion occurs they disappears or become in valid words difficult to identify and the introduction or error is possible. On this way, the number $F$ is high too, because the n-gram model can't offers information when an error appears in a low frequency n-gram, so when it is weighted by the module of selection of candidate for making the correction there is a low possibility to select the right one.

On the other hand, the PAEC method was affected by the POS-tagging in two aspects:

- In error detection: when a word is taken as an error but the POS-tag is not so precise, then it is searched in the wrong dictionaries; so, $I$ rate is increased.
- In potential correction generation: as in error detection, the candidates are generated from a possible wrong dictionary. So, the candidate can be different from the right one.

Finally, in **Table 6** can be seen the result from the POS-tagged files. As can be appreciated, the results are improved with respect to both previous tests. On this way, the PAEC proposal improves it performance given these facts:

- In error detection: now it is possible to identify wrong words that originates no valid words in Spanish, because can be matched in the right dictionary according to their POS-tag. Additionally, the errors that originate valid words are detected because the POS-tag indicate a low frequency n-gram: a wrong word belongs to a different grammatical category due to the mistake, and it is identified.
- In potential correction generation: the candidate generation is performed by its morphologic features, the n-gram probability and its grammatical category adding by this means the morphologic and syntactic knowledge.
- In correction: due to the correct detection the right candidate is selected in most of the times, so the correction is right.

## 5  Conclusions and Future Work

According to the results presented in the previous section the PAEC method is better in error correction when a precise POS-tagging is performed on the text because the morphologic and syntactic knowledge is properly added to the process, so the rationale is the right one but there are some improvement to do; if it is not the case, there are a lot of mistakes in the entire correction process.

Additionally, in order to improve the performance of the PAEC method it is required to do:

- To explore the use of syntactic n-grams (sn-grams), which are an improved form of n-grams, constructed using paths in syntactic trees in order to reduce the noise given to presence of many arbitrary terms in the n-gram. The sn-grams can be applied to the three stages of the error correction process to calculate POS sn-grams and word sn-grams [13].

- To improve the precision of the POS-tagger:
    - To train the algorithm in a more intensive way, adding more annotated text to star the learning of tagging rules.
    - To develop an alternative way to tagging words not identified and increase the lexical files of the tagger.
    - To use more than one tagger in order to improve the rate of correct tagging by the weighting the results of them.
    - To assign more than one POS-tag to words, accordingly to the probability of tags.
- To improve the detection of errors for searching the potential errors in more than one grammatical category lexicon given the assignation of more than one POS-tag.
- To refine the lexicons in two ways:
    - Increase the words they contain, in order to extend its lexical coverage.
    - By means of a lemmatization process, so the index of word would be more flexible because it can contains no only the complete orthographic form.

As a final commentary is the method can be tested on real world errors and a larger amount of texts. They can be taken from:
- Contents of Web 2.0 pages.
- Government pages of complains, administrative or legal processes and opinion pages.
- A bank of OCR-processed documents from libraries and government offices.